\documentclass[letterpaper, 10 pt, conference]{ieeeconf}  % Comment this line out if you need a4paper
\usepackage[pdftex]{graphicx}
\usepackage{color}
\usepackage{lipsum}
\usepackage{cite}
\usepackage{textcomp}
\usepackage{amsmath}
\DeclareGraphicsExtensions{.pdf,.png,.jpg}

\IEEEoverridecommandlockouts                              % This command is only needed if 
\title{\LARGE \bf
A Soft Magneto-Elastic Hysteresis Peristaltic Pump Toward Biomedical Flow Assistance
}

\author{Minjo Park$^{1}$ and Metin Sitti$^{2}$% <-this % stops a space
\thanks{*This work was not supported by any organization}% <-this % stops a space
\thanks{$^{1}$Minjo Park is with the Max Planck Institute for Intelligent System, 70569 Stuttgart, Germany
        {\tt\small minjo@is.mpg.de}}%
\thanks{$^{2}$Metin Sitti is a Professor in the School of Medicine and College of Engineering, Koç University,
        İstanbul,  Rumelifeneri Yolu 34450, Türkiye
        {\tt\small MSITTI@ku.edu.tr}}%
}

\begin{document}

\maketitle
\thispagestyle{empty}
\pagestyle{empty}

%%%%%%%%%%%%%%%%%%%%%%%%%%%%%%%%%%%%%%%%%%%%%%%%%%%%%%%%%%%%%%%%%%%%%%%%%%%%%%%%
\begin{abstract}

Pumping fluids plays a crucial role in various biomedical applications. Among different pumping mechanisms, peristalsis enables efficient and safe fluid transport by deforming a tube without direct contact with the fluid. While previous studies have proposed various actuation methods—such as mechanical, pneumatic, and magnetic actuation—to deform elastic membranes, these approaches often result in complex pump structures and control mechanisms. In this study, we propose a novel pump that generates peristaltic motion using a simple single pneumatic control combined with an embedded passive magnet. Computational simulations were conducted to predict the internal fluid flow, and the membrane motion observed in the simulations was validated through experiments with a physical prototype.

\end{abstract}

%%%%%%%%%%%%%%%%%%%%%%%%%%%%%%%%%%%%%%%%%%%%%%%%%%%%%%%%%%%%%%%%%%%%%%%%%%%%%%%%
\section{INTRODUCTION}
Pumping fluids plays a crucial role in nature, industry\cite{Bach2015}, and medical applications\cite{Maybaum2007}. In biological systems, fluid transport is achieved through the deformation of muscle tissues in various circulatory systems, enabling the efficient absorption and distribution of oxygen, blood, and nutrients. In biomedical fields, numerous pumping mechanisms have been developed for cardiac\cite{Han2019}, respiratory\cite{Simonds2006}, dialysis\cite{Groth2023}, and gastrointestinal\cite{smith2024} applications.

Among these mechanisms, peristaltic motion is observed in the digestive systems of animals, where localized deformations of a tube continuously transport internal fluids or solid-fluid mixtures in one direction \cite{Burns1967}. Inspired by this mechanism, mechanical tube-squeezing pumps have been commercialized\cite{Esser2019}, and various attempts have been made to generate artificial peristaltic motion using different actuation sources, such as hydraulic actuation\cite{Totuk2024} and magnetic fields\cite{Sharma2024}. However, these approaches often introduce complexity in both system structure and operation.

In this study, we propose a Magneto-Elastic Hysteresis Peristaltic Pump (MEHPP) that generates unidirectional net flow through a single pneumatic control input. The key components of this pump include an elastic membrane, permanent magnets attached to its surface, and a separated outer chamber for pressurization and depressurization. Owing to the difference between the restoring force of the elastic membrane and the interaction force between the magnets, hysteresis occurs during pressurization and depressurization, resulting in peristaltic motion. To analyze this principle, we conducted modeling and simulation, followed by the development of a simple proof-of-concept prototype. Given its structural simplicity\cite{Abhinav2025}\cite{Throckmorton2021}, this approach may offer a promising strategy for physiological biofluid systems in the human body, such as cardiovascular and lymphatic circulation, cerebrospinal fluid dynamics, urinary excretion, and pathological fluid accumulation, including ascites and pleural effusion.

\section{Modeling}

In our pump system, two primary forces are involved: external and internal forces. The external force is the pneumatic pressure applied to the elastic membrane, while the internal forces include the restoring force of the elastic membrane and the interaction force between the embedded permanent magnets. Additionally, the deformation of the tube in the z-direction is constrained by the internal and external contact boundaries, as shown in Fig.1.

\begin{figure}[htbp]
\centering
   \includegraphics[width=8cm]{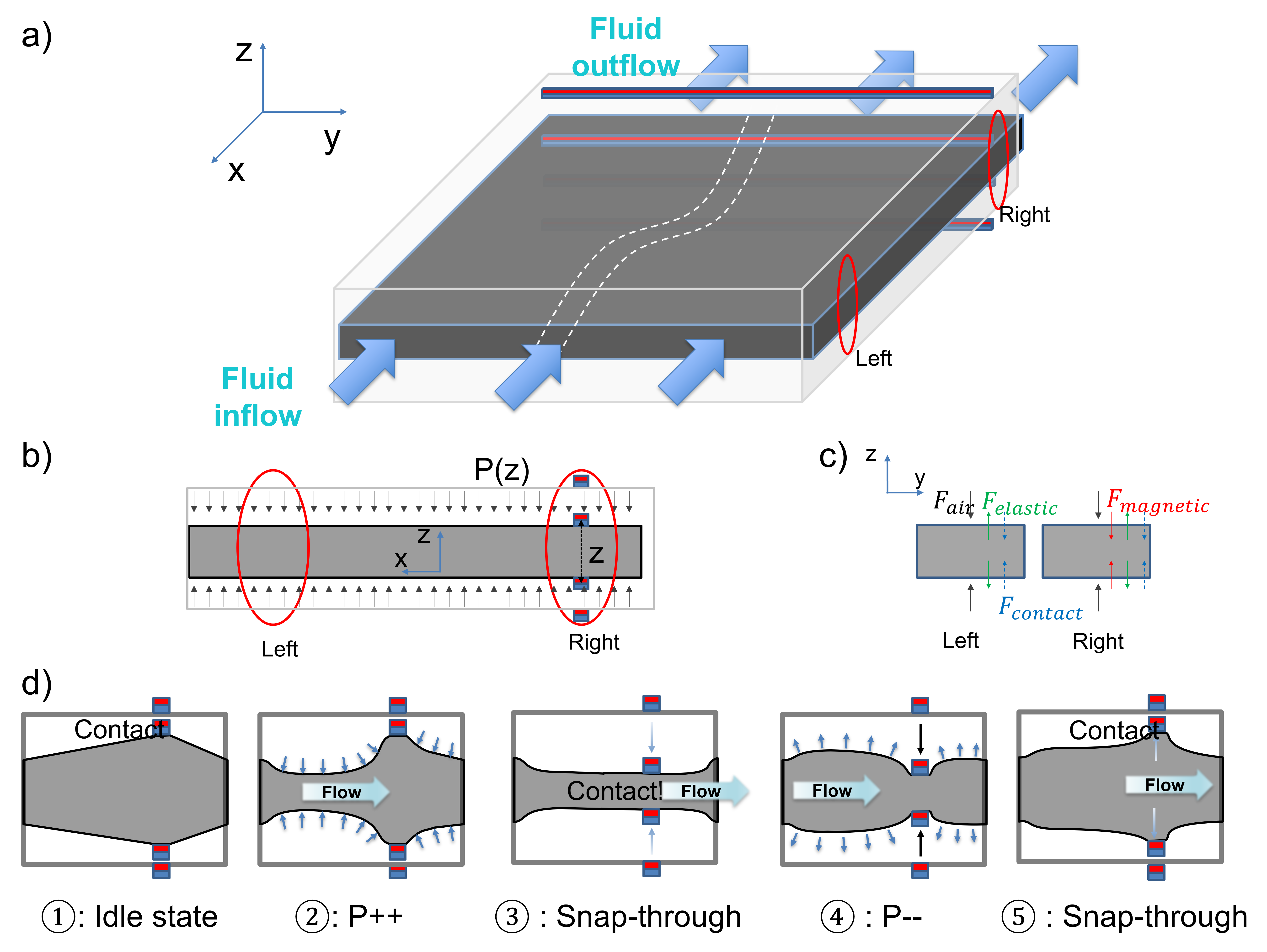}
   \hfil
\caption{Analytical model of the MEHPP. (a) 2D approximation and perspective view of the system. (b) Side view of the MEHPP. (c) Forces acting on the left and right parts of the MEHPP. (d) Schematic of the hysteresis-induced peristaltic sequence
}
\end{figure}

For a simplified mathematical modeling approach, we assumed a 2D approximation of the tube oriented along the x-axis, considering it to be infinitely extended in the y-direction, as shown in Fig.1. Additionally, we assumed that the restoring force of the elastic membrane acts linearly in the z-direction, while the interaction force between the magnets is inversely proportional to the cube of the distance. 
The displacement of the upper part of the elastic membrane from the origin is denoted as $z$. The natural length of the elastic membrane and the $z$-coordinates of the inner and outer boundaries are denoted as $z_0$, $z_{in}$, and $z_{out}$, respectively. Additionally, the spring constant of the elastic membrane, the magnetic force coefficient between the inner magnets, and the magnetic force coefficient between the inner and outer magnets are approximated as $k_e$, $k_{mi}$, and $k_{mo}$, respectively. Due to the boundary condition, $z$ satisfies the relation $z_{in} \leq z \leq z_{out}$. In the absence of contact between the elastic membrane and the boundary, the equilibrium pressure $p(z)$ can be derived from the equilibrium condition under quasi-static conditions.

$$p(z) = k_e (z_0 - z) + \frac{k_{mo}}{(z_1 - z)^3} - \frac{k_{mi}}{8z^3}$$

Additionally, the boundary positions are set by assuming $z_{out} = 1.25z_0$ and $z_{in} = 0.25z_0$. By dividing both sides by $k_e$ and $z_0$ to non-dimensionalize the equation, the following expression is derived:
$$p^* = 1 - z^* + \frac{a_{mo}}{(z^*_1 - z^*)^3} - \frac{a_{mi}}{8z^{*3}}$$

For an efficient and intuitive analysis, it is assumed that the sizes of the inner and outer magnets are equal, which implies that the magnetic force coefficients are also equal, i.e., $a_{mo} = a_{mi} = a$. Depending on the value of the $a$, the pressure profile versus z displacement changes. The critical value ($a_{crit}$) is 0.1341. 

If $ a $ exceeds a certain threshold, $ z $ does not exhibit local extrema within the range of 0 to 1.5 and instead continues to increase monotonically. However, if $ a $ is below this critical value, multiple extrema appear. The $ p^*-z $ graphs corresponding to various ranges of $ a $ are shown in Fig.2.
\begin{figure}[htbp]
\centering
   \includegraphics[width=7.5cm]{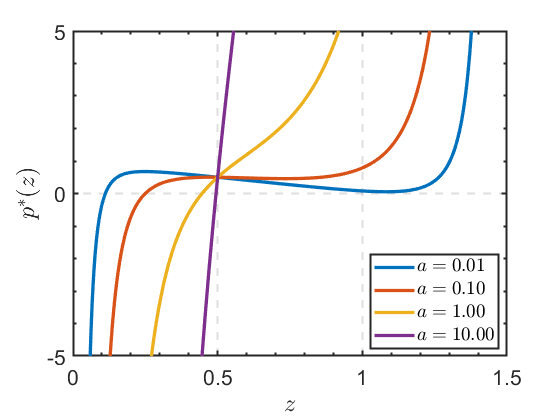}
   \hfil
\caption{ Profiles of $p^*(z)$ for differnent values of a.
}
\end{figure}

For cases where $ a=0.1 $, the trajectory in the $ p^*-z $ plane over a single cycle was analyzed, as shown in Fig.3. Also, the schematic of the actuation sequence is shown in Fig.1d.  Initially, when no pressure is applied, the system is at Position 1, which represents the open state of the conduit. In this state, a contact normal force is exerted from the external boundary. As pressure is applied, $ z $-displacement does not occur until this contact force is reduced to zero. After reaching Point 2, the system stabilizes at progressively lower pressures.

If the pneumatic pressure is alternated between two specific values, the state transitions from Point 2 to Point 3, moving parallel to the $ x $-axis. At Point 3, an internal normal contact force appears, maintaining contact as the system moves to Point 4. When negative pressure reaches the value corresponding to Point 4, the contact is released, allowing the system to transition along the $ x $-axis to Point 5, where contact with the outer boundary is reestablished. As pressure is reapplied, the system returns to Point 1, repeating the cycle.

\begin{figure}[htbp]
\centering
   \includegraphics[width=7.5cm]{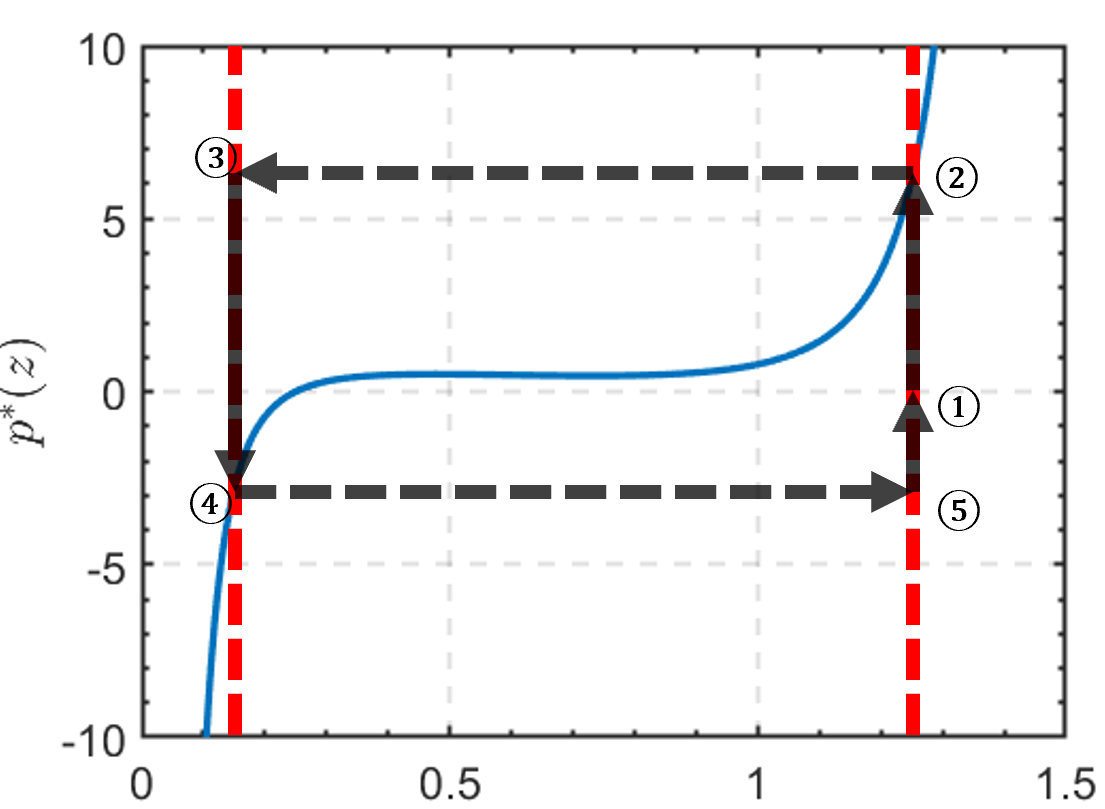}
   \hfil
\caption{Hysteresis loop of the deformation path over a pneumatic actuation cycle.
}
\end{figure}

During this process, the contact induced by the magnetic force plays a crucial role in generating hysteresis. If magnetic force were absent, each pressure value would correspond uniquely to a single $ z $-value, ensuring a one-to-one equilibrium condition without hysteresis. Consequently, the asymmetry in hysteresis behavior between the left and right sides of the pump leads to the emergence of one-directional flow.

By examining the cross-section in the xz plane, it is evident that magnets are absent on the left side and present on the right side. When pressure is applied, the membrane on the left side responds immediately, moving according to its equilibrium state. In contrast, the right side shows no displacement in the z-direction until the contact force between the membrane and the outer wall is reduced to zero. With further pressure increase, inner contact initiates on the left side and propagates rightward. This ultimately leads to a rapid transition via snap-through behavior, resulting in a squeezing motion directed to the right. Upon applying negative pressure, the left-side contact detaches immediately, while the right side maintains contact due to the strong magnetic force in the squeezing direction. As negative pressure increases, detachment propagates from left to right, drawing fluid from the left region into the tube. Once contact is released on the right, the tube returns to a fully inflated state. Reapplying positive pressure resets the system to its initial configuration, repeating the same sequence. In the absence of MEH, fluid is symmetrically pumped in both directions during pressurization and returns during depressurization, resulting in no net flow.

\section{Simulation}

MEH behavior and one-directional flow transportation were validated through COMSOL FSI module(version 6.2, COMSOL AB)simulations, where parameters were empirically tuned to maximize hysteresis, thereby generating a one-directional flow.

\begin{figure}[htbp]
\centering
   \includegraphics[width=8.5cm]{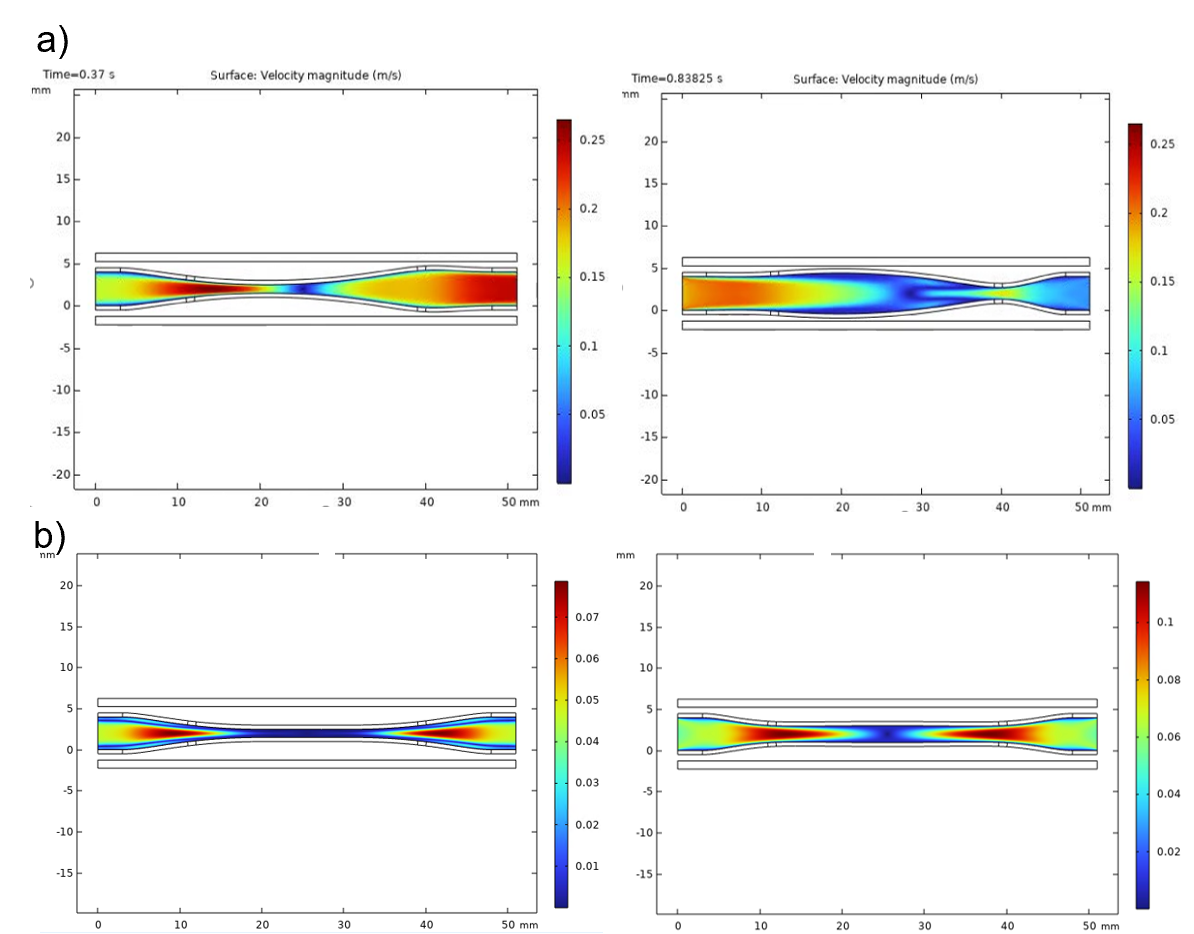}
   \hfil
\caption{Simulation results of the pneumatically actuated membrane pump at 0.37s and 0.83s: (a) with MEH, (b) without MEH.
}
\end{figure}

\begin{figure}[htbp]
\centering
   \includegraphics[width=7cm]{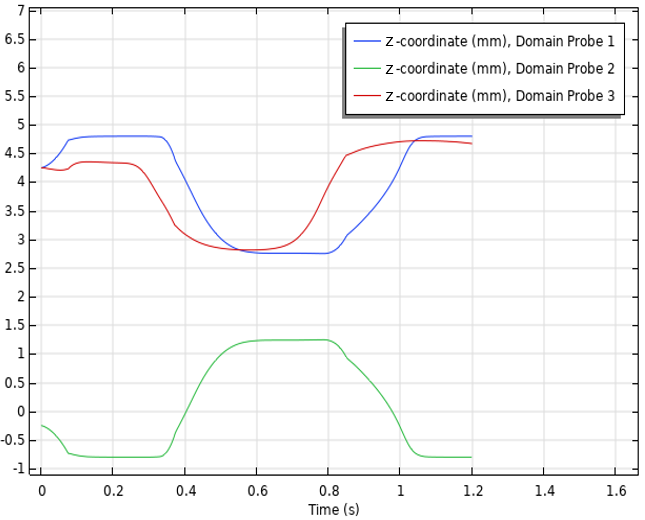}
   \hfil
\caption{Z-axis displacement versus time for different regions (Red—left upper, Blue—right upper, Green—right lower)
}
\end{figure}

We conducted simulations for two cases: with and without MEH. We set up a asymmetric configuration where the left side had no MEH while the right side included MEH, as shown in Fig.4. From 0 to 0.2 seconds, no pneumatic pressure is applied, and the system remains in equilibrium. Subsequently, a positive pressure is applied to the membrane from 0.3 to 0.7 seconds, while a negative pressure is applied from 0.5 to 1.1 seconds. Additionally, the simulation considers the occurrence of contact between the external boundary and the membrane, as well as between the internal boundaries. The z-direction coordinates at the center of each side were tracked using probes. The results, shown in Fig.5, indicate that over time, the right section closes and detaches later than the left section.
As a control case, we also conducted simulations for the scenario where MEH does not exist. As expected, the elastic membrane deformed symmetrically in the left-right direction, and the resulting internal fluid flow also exhibited perfect symmetry. This confirms that no net flow can exist in periodic operation under these conditions. 

\begin{figure}[htbp]
\centering
   \includegraphics[width=7.5cm]{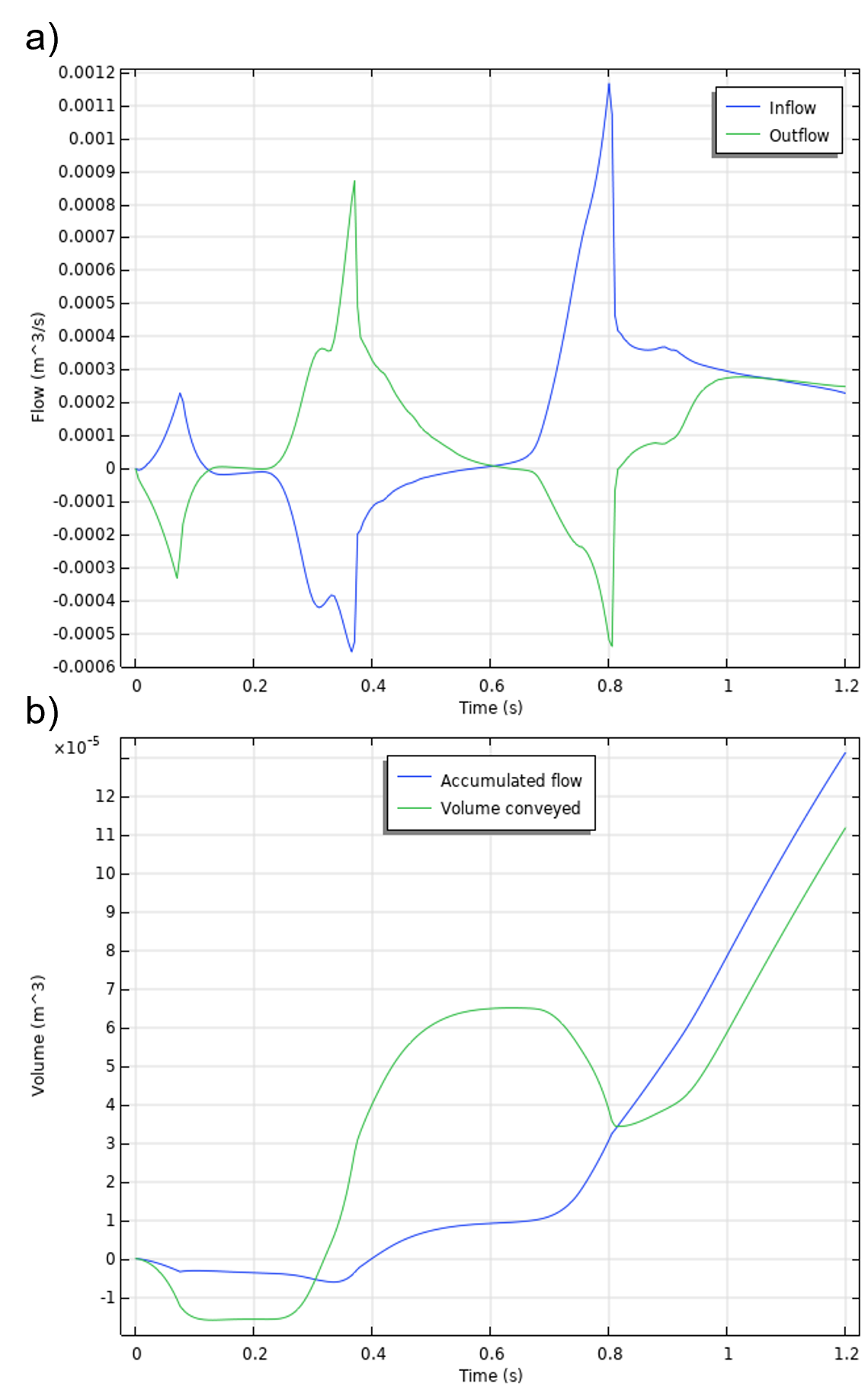}
   \hfil
\caption{Calculated flow properties. (a) Flow rates at the inlet (blue) and outlet (green). (b) Conveyed volume through the whole system (blue) and at the outlet (green).
}
\end{figure}

Additionally, we calculated the flow rate and the transferred volume of the pump. Figure 6a shows two curves of flow rate versus time at the inlet and outlet.
During the entire process, it is observed that the inflow value is negative during the pressurization phase, while the outflow value becomes negative during the depressurization phase. This occurs because, in the peristalsis process, it takes time for contact to form at one side and block the fluid flow. Additionally, to ensure the stability of the simulation, a contact boundary offset was introduced, preventing perfect physical contact as in reality, which results in slight leakage. Figure 6b illustrates the volume of fluid transferred by the actual pump in operation. Despite the occurrence of backflow, the forward flow in other regions is stronger, leading to a net flow. The two values in the graph are calculated as follows.
$$ \text{Accumulated flow}(t) = \int_{0}^{t} \frac{\text{Inflow} + \text{Outflow}}{2} \, dt $$
$$ \text{Volume conveyed}(t) = \int_{0}^{t} \text{Outflow} \, dt $$
However, the COMSOL FSI simulation was limited in reproducing the periodic behavior after one full cycle. After the first cycle, global one-directional flow and local vortices generated by the interaction with structural deformation caused a loss of convergence, leading to divergence in the simulation. This issue is expected to be addressable in future studies using alternative simulation tools and techniques.

\section{Proof-of-concept prototype and validation}

We fabricated a prototype of the proposed MEH peristaltic pump to verify its effectiveness in generating hysteresis and generating one-directional fluid flow.

\begin{figure}[htbp]
\centering
   \includegraphics[width=8.6cm]{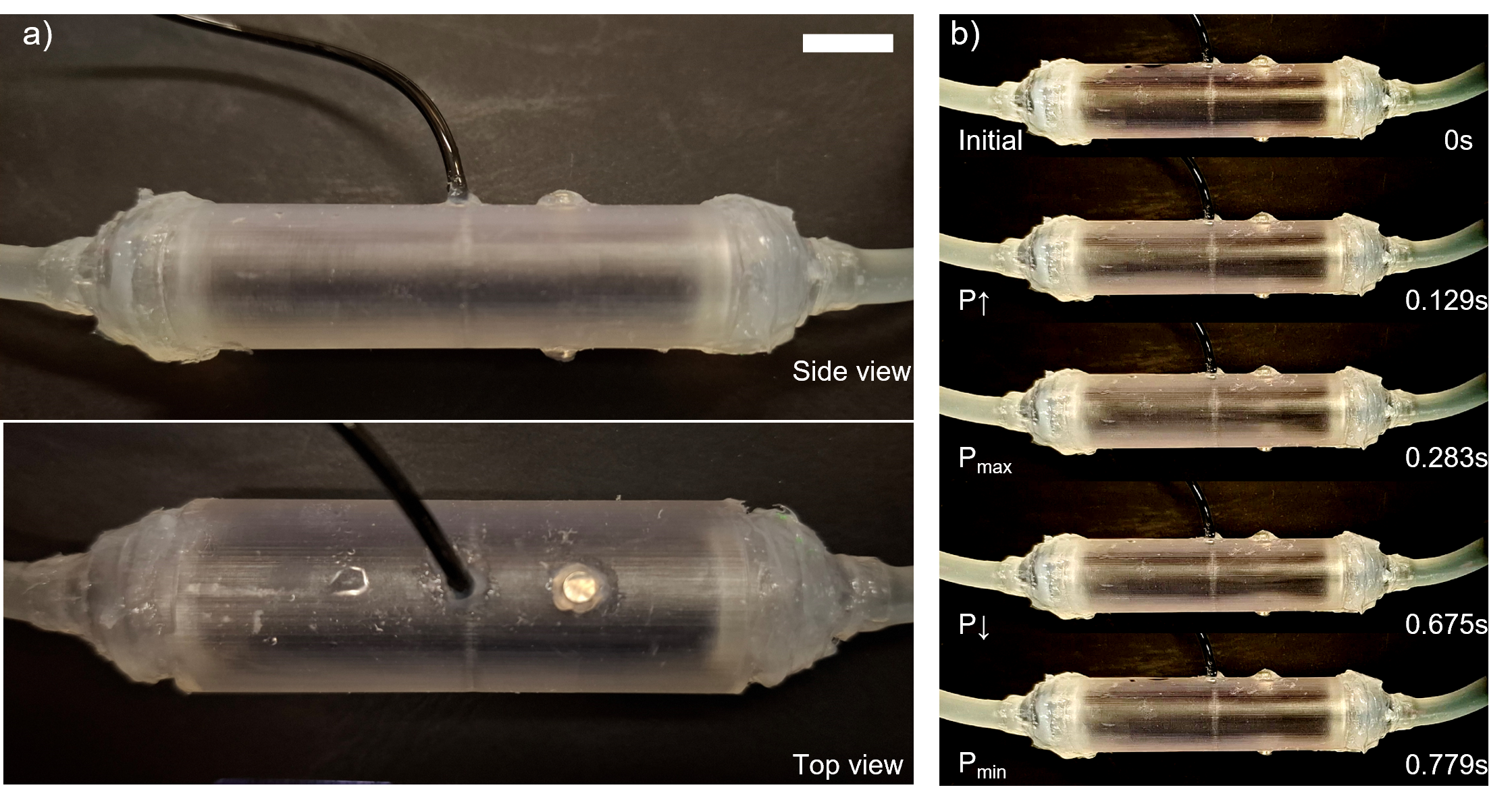}
   \hfil
\caption{MEHPP prototype. (a) Side and top views. (b) Actuation sequence of the MEHPP. Scale bar: 10mm
}
\end{figure}

The pump components were manufactured using 3D printing, including an elastic membrane (Silicone 40A resin) and the connection between the external rigid wall and the membrane (Clear V4, Formlabs). Each component was assembled using silicone adhesive (Silpoxy, Smooth-On) and instant adhesive (Loctite 401, Henkel). Finally, a urethane hose was connected for pneumatic control, and a valve and pump were integrated to apply periodic positive and negative pressures. The fabricated prototype is shown in Fig. 7, and the control system is illustrated in Fig. 8. A diaphragm pump, two pneumatic solenoid valves, and an Arduino were used to enable pressure switching every 500 ms.
\begin{figure}[htbp]
\centering
   \includegraphics[width=8.5cm]{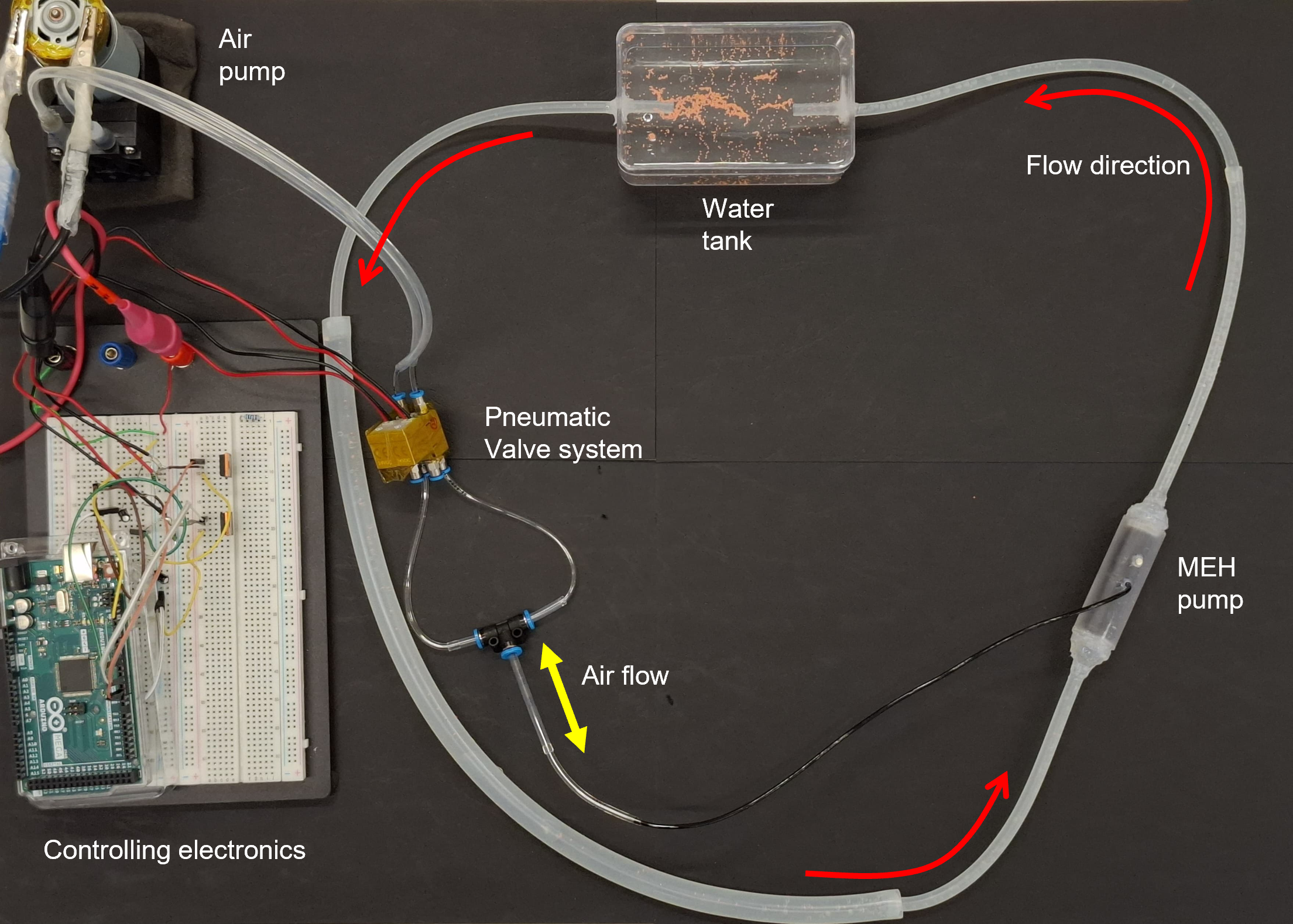}
   \hfil
\caption{Experimental setup consisting of the air pump, controller, and fluidic loop.
}
\end{figure}
Additionally, to effectively visualize the fluid flow inside the pump, floating particles (0.995 g/cc Red PE Microsphere, Cospheric) were mixed with water, and an air bubble region was introduced to facilitate an overall understanding of the flow dynamics. The experimental results are shown in Fig. 9. 

\begin{figure}[htbp]
\centering
   \includegraphics[width=8.5cm]{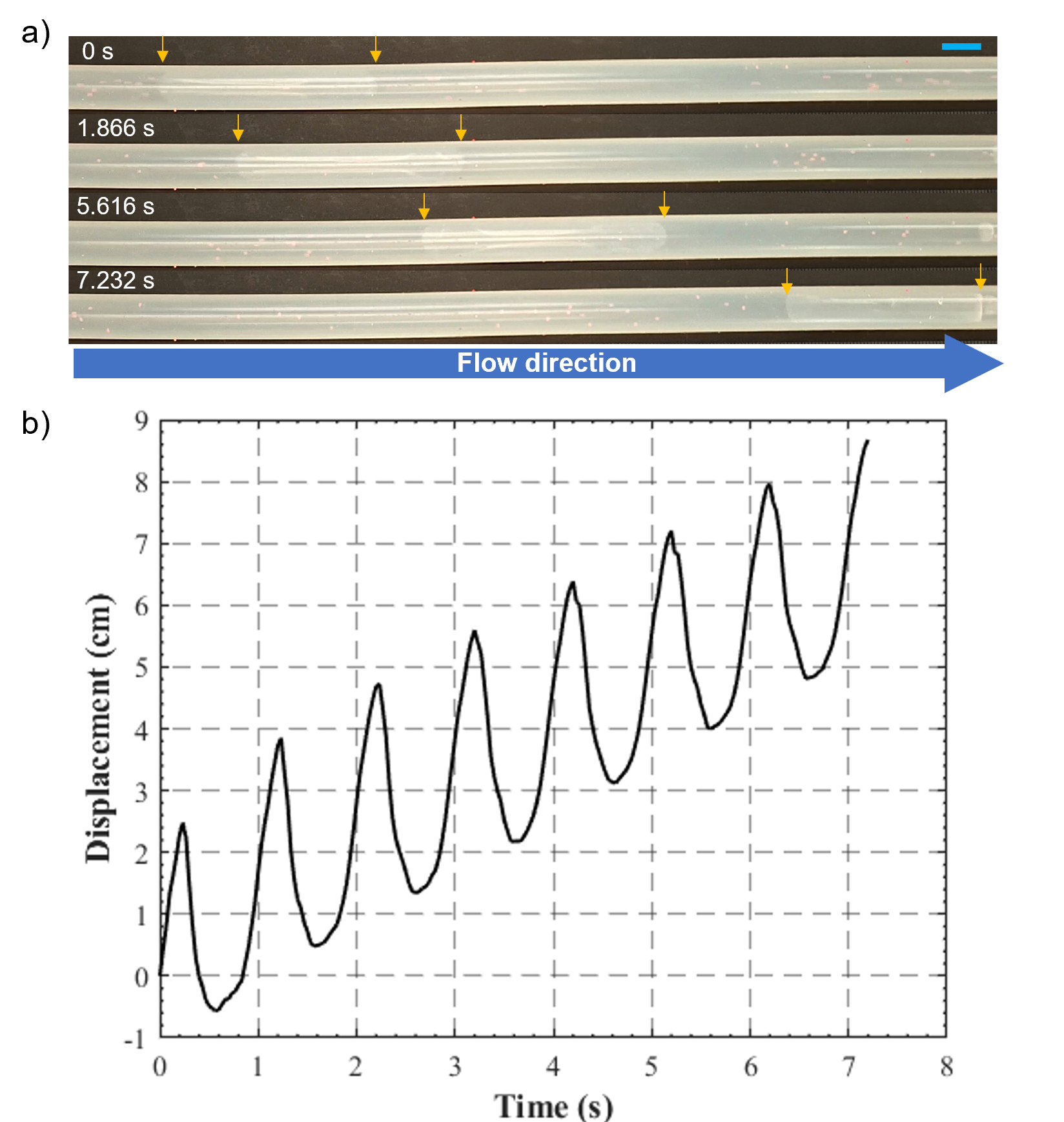}
   \hfil
\caption{Experimental validation of the MEHPP. (a) Fluid flow over time. (b) Flow-speed profile. Scale bar: 10mm.
}
\end{figure}
Due to the peristaltic motion induced by MEH, the air bubble continuously moves in one direction. Since the proposed prototype has a circular cross-section, perfect inner contact over a wide area, as assumed in the modeling, was not fully achieved, leading to slight backflow. However, this can be further improved by adopting a more efficient cross-sectional shape, such as a rectangular structure, which would enhance inner contact formation.

\section{Conclusion}
In this study, we proposed a highly simple pump structure incorporating an elastic membrane and permanent magnets, which is controlled using a single pneumatic control system. By employing magneto-elastic hysteresis, we successfully generated peristaltic motion and achieved unidirectional flow. This approach was supported by modeling and simulation, followed by the fabrication of a proof-of-concept prototype. The hysteresis-induced sequential deformation contributed to reducing backflow and improving pumping efficiency by promoting directional fluid transport. Since no additional valves or complex control elements are required, the system functions like a regular tube when pneumatic actuation is turned off. Notably, its minimalistic design may help reduce the risk of flow stasis, such as thrombosis, making it particularly advantageous for medical applications.

However, since this study analyzed a static situation without global flow, further validation under continuous or pulsatile flow conditions is required to ensure practical operation as an auxiliary device within physiological biofluid systems. Moreover, the unique characteristics of diverse biofluids—including their viscosity, non-Newtonian behavior, and suspended components, such as red blood cells or particulates—should be incorporated into future studies. These fluids may range from blood, urine, and cerebrospinal fluid to pathological fluids such as ascites and pleural effusion, and their consideration will be important for enhancing clinical relevance and broadening potential applications.

\addtolength{\textheight}{-12cm}   % This command serves to balance the column lengths

%%%%%%%%%%%%%%%%%%%%%%%%%%%%%%%%%%%%%%%%%%%%%%%%%%%%%%%%%%%%%%%%%%%%%%%%%%%%%%%%

%%%%%%%%%%%%%%%%%%%%%%%%%%%%%%%%%%%%%%%%%%%%%%%%%%%%%%%%%%%%%%%%%%%%%%%%%%%%%%%%

%%%%%%%%%%%%%%%%%%%%%%%%%%%%%%%%%%%%%%%%%%%%%%%%%%%%%%%%%%%%%%%%%%%%%%%%%%%%%%%%
\section*{ACKNOWLEDGMENT}
This work was supported by institutional funding from the Max Planck Society and the Max Planck Institute for Intelligent Systems.
\bibliographystyle{IEEEtran}
\bibliography{export}

@article{Totuk2024,
   author = {Onat Halis Totuk and Selçuk Mıstıkoğlu and Mehmet Ali Güvenç},
   doi = {10.1088/2631-8695/ad8ff6},
   issn = {2631-8695},
   issue = {4},
   journal = {Engineering Research Express},
   pages = {045232},
   title = {Design, optimization, simulation, and implementation of a 3D printed soft robotic peristaltic pump},
   volume = {6},
   url = {},
   year = {2024}
}

@article{Sharma2024,
   author = {Saksham Sharma and Laura Caroline Jung and Nicholas Lee and Yusheng Wang and Ane Kirk-Jadric and Rishi Naik and Xiaoguang Dong},
   doi = {10.1002/adfm.202405865},
   issn = {16163028},
   journal = {Advanced Functional Materials},
   publisher = {John Wiley and Sons Inc},
   title = {Wireless Peristaltic Pump for Transporting Viscous Fluids and Solid Cargos in Confined Spaces},
   year = {2024}
}

@misc{Bach2015,
   author = {D. Bach and F. Schmich and T. Masselter and T. Speck},
   doi = {10.1088/1748-3190/10/5/051001},
   issn = {17483190},
   issue = {5},
   journal = {Bioinspiration and Biomimetics},
   pmid = {26335744},
   publisher = {Institute of Physics Publishing},
   title = {A review of selected pumping systems in nature and engineering - Potential biomimetic concepts for improving displacement pumps and pulsation damping},
   volume = {10},
   year = {2015}
}

@article{Maybaum2007,
   author = {Simon Maybaum and Donna Mancini and Steve Xydas and Randall C. Starling and Keith Aaronson and Francis D. Pagani and Leslie W. Miller and Kenneth Margulies and Susan McRee and O. H. Frazier and Guillermo Torre-Amione},
   doi = {10.1161/CIRCULATIONAHA.106.633180},
   issn = {00097322},
   issue = {19},
   journal = {Circulation},
   pages = {2497-2505},
   pmid = {17485581},
   title = {Cardiac improvement during mechanical circulatory support: A prospective multicenter study of the LVAD working group},
   volume = {115},
   year = {2007}
}

@misc{Han2019,
   author = {Jooli Han and Dennis R. Trumble},
   doi = {10.3390/bioengineering6010018},
   issn = {23065354},
   issue = {1},
   journal = {Bioengineering},
   publisher = {MDPI AG},
   title = {Cardiac assist devices: Early concepts, current technologies, and future innovations},
   volume = {6},
   year = {2019}
}

@article{Simonds2006,
   author = {Anita K. Simonds},
   doi = {10.1378/chest.130.6.1879},
   issn = {00123692},
   issue = {6},
   journal = {Chest},
   pages = {1879-1886},
   pmid = {17167012},
   publisher = {American College of Chest Physicians},
   title = {Recent advances in respiratory care for neuromuscular disease},
   volume = {130},
   year = {2006}
}

@misc{Groth2023,
   author = {Thomas Groth and Bernd G. Stegmayr and Stephen R. Ash and Janna Kuchinka and Fokko P. Wieringa and William H. Fissell and Shuvo Roy},
   doi = {10.1111/aor.14396},
   issn = {15251594},
   issue = {4},
   journal = {Artificial Organs},
   pages = {649-666},
   pmid = {36129158},
   publisher = {John Wiley and Sons Inc},
   title = {Wearable and implantable artificial kidney devices for end-stage kidney disease treatment: Current status and review},
   volume = {47},
   year = {2023}
}

@article{Burns1967,
   author = {J. C. Burns and T. Parkes},
   doi = {10.1017/S0022112067001156},
   issn = {14697645},
   issue = {4},
   journal = {Journal of Fluid Mechanics},
   pages = {731-743},
   title = {Peristaltic motion},
   volume = {29},
   year = {1967}
}

@article{Esser2019,
   author = {Falk Esser and Tom Masselter and Thomas Speck},
   doi = {10.1002/aisy.201900009},
   issn = {2640-4567},
   issue = {2},
   journal = {Advanced Intelligent Systems},
   publisher = {Wiley},
   title = {Silent Pumpers: A Comparative Topical Overview of the Peristaltic Pumping Principle in Living Nature, Engineering, and Biomimetics},
   volume = {1},
   year = {2019}
}

@article{Throckmorton2021,
   author = {Amy Throckmorton and Ellen Garven and Matthew Hirschhorn and Steven Day and Randy Stevens and Vakhtang Tchantchaleishvili},
   doi = {10.21037/acs-2020-cfmcs-14},
   issn = {23041021},
   issue = {3},
   journal = {Annals of Cardiothoracic Surgery},
   pages = {383-385},
   publisher = {AME Publishing Company},
   title = {Forward-thinking design solutions for mechanical circulatory support: multifunctional hybrid continuous-flow ventricular assist device technology},
   volume = {10},
   year = {2021}
}

@misc{Abhinav2025,
   author = {Vishnuram Abhinav and Prithvi Basu and Shikha Supriya Verma and Jyoti Verma and Atanu Das and Savita Kumari and Prateek Ranjan Yadav and Vibhor Kumar},
   doi = {10.3390/mi16050522},
   issn = {2072666X},
   issue = {5},
   journal = {Micromachines},
   publisher = {Multidisciplinary Digital Publishing Institute (MDPI)},
   title = {Advancements in Wearable and Implantable BioMEMS Devices: Transforming Healthcare Through Technology},
   volume = {16},
   year = {2025}
}

@article{smith2024,
  title   = {MagnetoStalsis: Generating Peristalsis in an Artificial Bowel for Treatment of Short Bowel Syndrome},
  author  = {Smith, Mariana E. and May, Adam and Schwehr, Trevor and Erin, Onder and Tragesser, Cody and Scheese, Daniel and Mair, Lamar O. and Diaz-Mercado, Yancy and Hackam, David and Krieger, Axel},
  journal = {Journal of Medical Robotics Research},
  volume  = {9},
  number  = {03n04},
  pages   = {2440006},
  year    = {2024},
  doi     = {10.1142/S2424905X24400063}
}

\end{document}